\documentclass[letterpaper, 10 pt, conference]{ieeeconf}  % Comment this line out if you need a4paper
\makeatletter
\let\NAT@parse\undefined
\makeatother

\IEEEoverridecommandlockouts                              % This command is only needed if 
\usepackage{graphics} % for pdf, bitmapped graphics files
\usepackage{epsfig} % for postscript graphics files

\usepackage{amsmath} % assumes amsmath package installed
\usepackage{amssymb}  % assumes amsmath package installed
\usepackage{multicol}
\usepackage{multirow}
\usepackage{tabu}
\usepackage{booktabs}
\usepackage{subcaption}
\usepackage{bm}
\usepackage[bookmarks=true]{hyperref}
\usepackage{graphicx}
\usepackage{xcolor}
\usepackage{bbm}
\definecolor{Blue}{rgb}{0,0,1}
\usepackage[raggedrightboxes]{ragged2e}

\newcommand{\todocite}[1]{\textcolor{Blue}{[Citation needed]}}

\newcommand{\bI}{{\bf I}} % rgbd image
\newcommand{\bC}{{\bf C}} % color image
\newcommand{\bD}{{\bf D}} % depth
\newcommand{\bT}{{\bf T}} % pose
\newcommand{\bS}{{\bf S}} % semantic

\newcommand{\bV}{{\bf V}} % volume
\newcommand{\bv}{{\bf v}} % voxel
\newcommand{\bs}{{\bf s}} % semantic

\usepackage{subcaption}
\usepackage{dsfont}
\usepackage
[backend=bibtex,
bibstyle=ieee,
citestyle=numeric,
sortcites,
natbib=true,
doi=false,
isbn=false,
url=false,
hyperref=true,
sorting=nyt,
eprint=false,
maxbibnames=99]{biblatex}
\bibliography{joaoreferences, otherreferences}

\title{\LARGE\bf On the Overconfidence Problem in Semantic 3D Mapping
}

\author{Joao Marcos Correia Marques$^1$, Albert Zhai$^1$, Shenlong Wang$^1$, and Kris Hauser$^1$% <-this % stops a space
\thanks{J. Marques was supported by NIFA Award \#2021-67021-34418.}% <-this % stops a space
\thanks{$^1$: Department of Computer Science at the University of Illinois at Urbana-Champaign, Urbana, Illinois, USA.      
        {\tt\small (jmc12,azhai2,shenlong,kkhauser)\@illinois.edu}}%
}

\begin{document}

\maketitle
\thispagestyle{empty}
\pagestyle{empty}
\thispagestyle{plain}
\pagestyle{plain}

% \input{macros}

%%%%%%%%%%%%%%%%%%%%%%%%%%%%%%%%%%%%%%%%%%%%%%%%%%%%%%%%%%%%%%%%%%%%%%%%%%%%%%%%
\begin{abstract}

Semantic 3D mapping, the process of fusing depth and image segmentation information between multiple views to build 3D maps annotated with object classes in real-time, is a recent topic of interest. This paper highlights the fusion overconfidence problem, in which conventional mapping methods assign high confidence to the entire map even when they are incorrect, leading to miscalibrated outputs. Several methods to improve uncertainty calibration at different stages in the fusion pipeline are presented and compared on the ScanNet dataset.  We show that the most widely used Bayesian fusion strategy is among the worst calibrated, and propose a learned pipeline that combines fusion and calibration, GLFS, which achieves simultaneously higher accuracy and 3D map calibration while retaining real-time capability. We further illustrate the importance of map calibration on a downstream task by showing that incorporating proper semantic fusion on a modular ObjectNav agent improves its success rates. Our code will be provided on Github for reproducibility upon acceptance.

\end{abstract}

%%%%%%%%%%%%%%%%%%%%%%%%%%%%%%%%%%%%%%%%%%%%%%%%%%%%%%%%%%%%%%%%%%%%%%%%%%%%%%%%

\section{INTRODUCTION}
\vspace{-5px}
Confidence calibration captures the concept of the disparity between a given model's confidence in its output and its actual measured performance and has been widely studied in 2D vision tasks \cite{pmlr-v70-guo17a,9130729,kuppers2022confidence,Zhang2021DenseEstimation,ABASPURKAZEROUNI2022117734}. Calibrated uncertainty estimates can be crucial to decision making in many fields, such as autonomous driving \cite{Peng2021UncertaintyVehicles,NIPS2017_2650d608} and medical imaging \cite{9130729,9434131,MURUGESAN2023102826}, as it allows for more cautious action under uncertain conditions, improving safety and reliability. Semantic 3D mapping is an active research topic that seeks methods to build 3D maps in real-time while also leveraging image segmentation models from computer vision~\cite{7989538}, and has numerous applications in robotics.  However, the confidence calibration of its 3D maps has yet to be addressed. %despite its implications to autonomous active search and rescue planning and mapping \cite{asgharivaskasi2021active}.

%Existing confidence calibration literature either  works on finalized 3D representations \cite{9735278,Yeung2023CalibratingSegmentation}, which are ill-suited for the incremental real-time nature of metric-semantic mapping
%KH: those references are on on 3D medical image segmentation
%or focus on model calibration at the pixel space\cite{Peng2021UncertaintyVehicles,NEURIPS2019_f8c0c968,gupta2022toplabel,NEURIPS2019_8ca01ea9,9130729}. %KH: Already talked about that above.

\begin{figure}[tbp]
  \centering
    \includegraphics[width=0.99\linewidth]{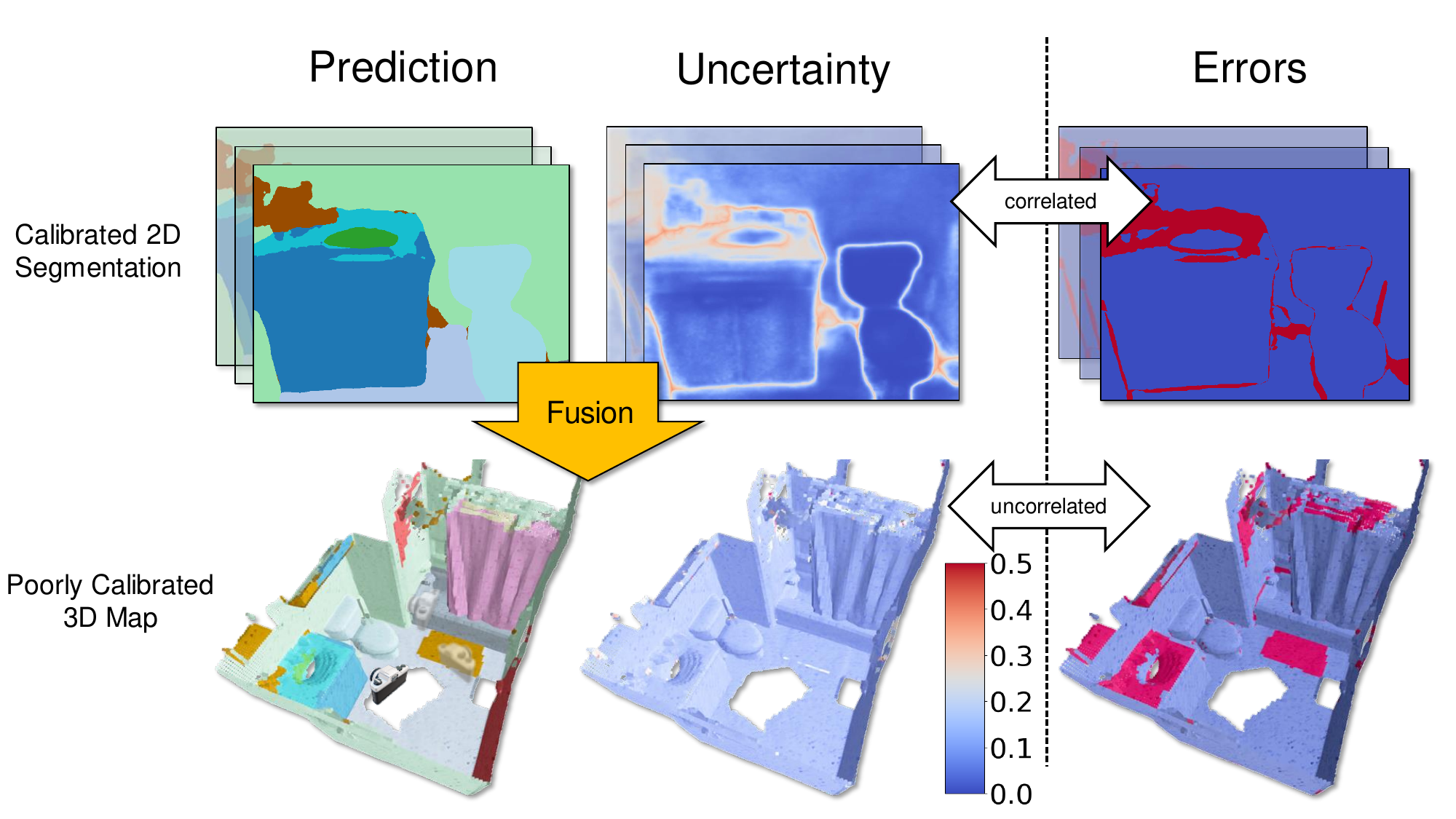}
    \caption{The problem of fusion overconfidence: In practice, standard Recursive Bayesian semantic fusion~\cite{7989538} produces overconfident 3D semantic maps with low uncertainty (bottom) in error regions even when the 2D semantic segmentation images are well-calibrated (with uncertainty correlating to errors, see top). [Best viewed in color.]
    % Recursive Bayesian Update Fusion \cite{7989538}
    }
    \label{fig:Fusion Overconfidence Problem}
    \vspace{-20px}
\end{figure}

In this paper, we articulate a general {\bf overconfidence problem} in semantic fusion calibration in which the semantic estimates in the 3D map become highly overconfident (Fig.~\ref{fig:Fusion Overconfidence Problem}).  Moreover, we highlight that overconfidence persists even when the image segmentation model is well calibrated (\autoref{tab: 2D calibration is insufficient}). Possible causes of this phenomenon include independence assumptions used in the fusion strategy, sensitivity to outliers, and viewpoint distribution biases. 
Moreover, we show the most widely used fusion strategy, Recursive Bayesian Update (RBU) \cite{7989538}, to be especially susceptible to overconfidence. 

We identify several approaches to mitigate these problems and show that adopting alternative fusion strategies, like Na\"ive Averaging \cite{8490953}, tend to improve calibration.  We introduce a method for calibrating the image segmentation model via Temperature / Vector Scaling~\cite{pmlr-v70-guo17a} to directly optimize the calibration of final fused maps. Moreover, we present Generalized Learned Fusion Strategy (GLFS), an end-to-end trainable framework that learns logit Vector Scaling, sample weighing, and fusion method in a unified fashion, while remaining real-time capable. Experiments on the ScanNet dataset show that these strategies improve map calibration without degrading accuracy. 
Finally, we show the utility of better calibrated maps in a robotics application. In an object-goal-navigation task~\cite{chaplot2020learning}, better calibrated maps lead to performance improvements as the agent can better handle overconfident segmentation model outputs through leveraging diverse object viewpoints for reliable detection. 

% \begin{figure}[tbp]
%   \centering
%     \includegraphics[width=0.90\linewidth]{figures/3D Calibration Results/joao_teaser.pdf}
%     \caption{The fusion overconfidence problem: in practice, standard Bayesian semantic fusion~\cite{7989538} yields overconfident 3D semantic maps (bottom) even if the 2D semantic segmentations are well-calibrated (top).
%     % Recursive Bayesian Update Fusion \cite{7989538}
%     }
%     \label{fig:Fusion Overconfidence Problem}
%     \vspace{-20px}
% \end{figure}

\section{RELATED WORK}
\vspace{-5px}
Semantic mapping aims to create a labelled 3D map of the environment from a sequence of RGB-D images or depth readings. Individual frames are given to semantic segmentation models to produce pixel labels that are fused into estimates for each map element. Methods vary in their choice of map representation, such as voxels \cite{rosinol2021kimera,Hughes2022Hydra:Optimization,9981300,asgharivaskasi2021active,8490953}, surfels \cite{7989538,8613746}, points \cite{8206392,6907236}, Neural Radiance Fields \cite{Kundu_2022_CVPR} or sparse Gaussian Processes \cite{9341658}, and some methods identify objects as sub-elements of the map~\cite{8206392,8613746,8741085}.   The overconfidence problem generally exists across map representations. Methods also vary in the choice of SLAM back-end, semantic segmentation model, and fusion strategy.  Existing fusion strategies include Recursive Bayesian Update (RBU) \cite{BULTMANN2023104286,7989538,8967890,rosinol2021kimera,Hughes2022Hydra:Optimization,9981300,asgharivaskasi2021active,Morilla-Cabello2023RobustMapping}, histogram/weighed average-based strategies \cite{8741085,chaplot2020learning,8490953,7138983,9811877} and learning-based aggregation \cite{Kundu_2022_CVPR,Fu2022PanopticSegmentation,xiang2017rnn,9341658}. Although most past work focuses on reconstruction accuracy or efficiency, overconfidence has also been noted~\cite{8490953,Morilla-Cabello2023RobustMapping}. It has been addressed via assorted methods such as using averaging fusion ~\cite{8490953} and sample weighing via epistemic uncertainty from Bayesian neural networks~\cite{Morilla-Cabello2023RobustMapping}, but a systematic study of semantic map calibration, as we do here, has not yet been done.

%These methods focus on improving the reconstruction accuracy or efficiency, but none of them studies the confidence calibration of the resulting map, which we do in this work.% - apart from  \citet{8490953}, who claim that heuristically averaging pixel probabilities yielded less overconfident 3D maps, but provided no evidence to back the claim. 

%Real time metric-semantic mapping pipelines aim to construct a semantically labelled representation of an environment given a series of observations in real time, extracting semantic information from deep networks. 
%The real time-constraints impose limits the techniques that can be applied to these reconstructions. \citet{7989538}

%Core Citations:
 % \cite{7989538}, \cite{8490953}, \cite{9981300}, \cite{rosinol2021kimera}, \cite{Hughes2022Hydra:Optimization},\cite{8613746}, \cite{6907236},\cite{8206392}\cite{Kundu_2022_CVPR}, \cite{jatavallabhula2023conceptfusion},\cite{kerr2023lerf},\cite{Morilla-Cabello2023RobustMapping}

 % Divide them by:
 % \begin{itemize}
 %     \item spatial representation (voxels, points, surfels, implicit)
 %     \item Fusion Strategy: Naive, Averaging, Histogram
 % \end{itemize}
Calibration has been an important topic in deep learning due to observed model overconfidence using standard training methods \cite{NEURIPS2019_f8c0c968,pmlr-v70-guo17a,xiong2023hingewasserstein}.  Model calibration is most often assessed using reliability diagrams \cite{10.1145/1102351.1102430}, which relates a model's confidence to the likelihood that it is correct. The most widely used metric is Expected Calibration Error (ECE)~\cite{naeini2015obtaining,pmlr-v70-guo17a}, and a variety of post-hoc calibration methods have been used to recalibrate models including Temperature and Vector Scaling\cite{pmlr-v70-guo17a} and Platt scaling \cite{platt1999probabilistic}, with many more extensions aimed at better calibrating multi-class predictors \cite{NEURIPS2019_8ca01ea9,gupta2022toplabel} or Detectors \cite{kuppers2022confidence}. The literature has also identified many issues with ECE, in particular for multi-class classification~\cite{Nixon2019MeasuringLearning,gupta2022toplabel}, like class frequency dependence, sensitivity to bin number and non-differentiability \cite{Bohdal2021Meta-Calibration:Error}.  Calibration has not yet been applied to semantic fusion because the ``model'' in this case is the output of a (potentially large) set of image segmentation model outputs fused into a segmentation of a 3D map element (voxel).  This paper proposes to measure and optimize the calibration of an entire semantic fusion pipeline, which represents a challenge across all views and voxels.
%\citet{kuppers2022confidence} Doesn't add much
%\citet{NEURIPS2019_f8c0c968} Not really relevant
%\citet{NEURIPS2021_8420d359} Shows that more modern learning models are better calibrated 
%\citet{9434131} \citet{9130729}  application to medical images
%\citet{pmlr-v162-marx22a}
%\citet{Ohno2006One-pointMethod}

%Explain the core idea behind confidence calibration, how it is measured (ECE, issues with it, top level CE and mECE)  and how it is improved (histogram binning, vector and temperature scaling, ensembling)

%Explain why 3D online calibration is different from all previous approaches (final output is not the raw model's, but the result of the fusion from several different observations), explain challenges and complexities associated with it (maybe include a teaser for our 2D != 3D results)?

Maps with calibrated uncertainty are helpful in several applications.  Past work has used semantic uncertainty as an objective for next-best-view problems for improved semantic reconstruction ~\cite{engelmannopen} or as information gain objectives for active map exploration~\cite{asgharivaskasi2021active}.
We apply this to object goal navigation (ObjectNav) where an agent, placed in an unknown environment, navigates to a specific object (e.g., "chair") using posed RGB-D observations~\cite{batra2020objectnav}.
We show that improved calibration in semantic fusion increases the success rates of a recent ObjectNav agent~\cite{zhai2023peanut} while using the same base segmentation model for semantics.

\section{BACKGROUND}

% Semantic 3D mapping incrementally builds an environment representation from a roving camera while simultaneously annotating the map position, color, and semantics.  For real-time use, the map is represented as a voxel grid with each cell encoding occupancy, aspects of internal geometry (such as distance or best fit plane), color, and semantic label.  Semantic annotations estimate the type of object contained in a voxel, and are expressed as membership into one of $M$ known classes (possibly including an ``other'' category).  Metric and color information is derived from the depth and RGB channels of the camera, respectively,  while semantic information is mostly derived from semantic or panoptic image segmentation models \cite{https://doi.org/10.48550/arxiv.2105.15203,He_2017_ICCV,https://doi.org/10.48550/arxiv.2105.15203}. The process of accumulating voxel annotations over time, merging the information from several viewpoints, is called \textbf{fusion}.  

We begin with a brief summary of semantic 3D mapping, %. While various forms of representation exist~\cite{6599048, doi:10.1177/0278364916669237}, this paper focuses
focusing on voxel representations~\cite{curless1996volumetric, 9918017,niessner2013real}. %, which is the most commonly adopted parameterization for real-time semantic 3D mapping.
Each entity in the voxel grid encodes geometry \cite{niessner2013real} (such as occupancy or signed distance field), appearance (e.g., colors), and semantic or instance labels~\cite{https://doi.org/10.48550/arxiv.2105.15203,He_2017_ICCV,https://doi.org/10.48550/arxiv.2105.15203} (in the form of a categorical probability vector). Given a series of sensor observations, the {\bf fusion} procedure aggregates the information from different viewpoints over time into each voxel. 

\subsection{Real-time Metric Reconstruction}

For metric mapping we revisit TSDF-based fusion~\cite{9918017,niessner2013real}. The mapping algorithm takes as input a series of calibrated, posed RGB-D images $\bI^t = \{(\bC^t ,\bD^t, \bT^t)\}_{t=1,\ldots, T}$, where $\bC^t \in \mathbb{R}^{H\times W \times 3}$ is the RGB color, $\bD^t  \in \mathbb{R}^{H\times W}$ is the depth and $\bT^t \in SE(3)$ is the pose generated by SLAM.  
The output of the metric reconstruction is a voxel grid $\bV = \{\bv_i = (\delta_i, w_i, \boldsymbol{\gamma}_i)\}_{i = 1,\ldots N}$ that encodes the geometry of 3D space. Each voxel  $\bv_i$ represents geometry attributes at a 3D location and contains three values: a signed distance to the nearest surface $\delta_i$, a weight $w_i$ used for averaging, and, optionally, a color tuple $\gamma_i$. 

At each step of the reconstruction stage $t$, the voxels within the camera frustum are projected into image space, associating each visible voxel $\bv_i$ with an image coordinate using $\bT^t$ and the camera's intrinsic parameter. Specifically, incremental weighted averaging fuses the new observation at time $t$ to update the values of each voxel as $
    \delta_i \gets \frac{w_i\delta_i + w_i^t\delta_i^t}{w_i^t+w_i}, 
    \mathbf{\gamma}_i \gets \frac{w_i\mathbf{\gamma}_i + w_i^t c_i^t}{w_i^t+w_i}, 
    w_i \gets w_i + w_i^t $,
where $\delta_i^t$ is the new observed signed distance function computed by taking the difference between the observed depth from the image and the projected depth of the voxel: $\delta_i^t = d_i^t - \hat{d}_i^t$; $c_i^t$ is the observed color of $\bv_i$ from $\bC^t$ and $w_i^t$ is the fusing weight.
%Observed depth $d_i^t$ and color $c_i^t$ of voxel $i$ are queried from $\bI^t$ by projecting voxel center $\mathbf{x}_i$ onto the image coordinate using $\bT^t$ and the camera's intrinsic parameters.
Only voxels visible within the truncation distance are updated to account for unknown areas behind the surface. The reconstructed surface mesh can be obtained using the marching cubes algorithm~\cite{1361981470015626752}.

\subsection{Semantic Fusion}
\label{sec:Fusion Strategies described}

Given the geometric representation, {\bf semantic fusion} additionally assigns a semantic label $l_i$ to each voxel $\bv_i$. To achieve this, semantic information is extracted from 2D image observations across viewpoints and integrated over visible voxels.  A class probability vector $\bs_i \in \mathbb{R}^K$ is associated with the voxel $\bv_i$ where the k-th element represents the probability that the voxel belongs to the semantic category $k$: $s_{ik} = P(l_i = k | \bI_{1,\dots, T})$, ensuring $\sum_k s_{ik} = 1$.  Each incoming image $\bI^t$ is segmented to generate a semantic map $\bS^t \in \mathbb{R}^{W\times H \times K}$. Then, for each voxel $\bv_i$ visible on frame $t$, we project it into image coordinates and query its semantic class vector $s_{ik}^t$ (accounting for visibility). This value is designated as the likelihood probability: $P(l_i = k | \bI^t) = \bs_i^t$. We wish to model the posterior of the voxel label given these observations: $P(l_i = k | \bI_{1,\dots, T})$.  

By assuming independence between observations, \citet{7989538} propose what is known as the \textbf{Recursive Bayesian Update} (RBU), which approximates the posterior through Bayes's Rule as:

\begin{equation}
\label{eq:bayes_integration}
\bs_i = P(l_i | \bI_{1,\dots, T}) \propto P(l_i) \prod_{t\in F_i} \bs_i^t, 
\vspace{-5px}
\end{equation}
$P(l_i)$ representing a prior distribution and $F_i$ the set of frames in which $\bv_i$ is observed.  Multiplication is performed element-wise. Assuming a uniform prior, $\mathbf{s}_i = \frac{1}{Z} \prod_t \mathbf{s}_i^t$,
where $Z$ normalizes $\bs_i$ into a valid probability vector. This approximation can be recursively estimated via $\bs_i \gets \frac{1}{Z} \bs_i\bs_i^t$.  
%To attain real-time efficiency and utilize GPU parallelization, this implementation of semantic fusion is \textbf{incremental} in nature. In particular, the \textbf{Bayesian integration} strategy~\todocite assumes the observations are sequential. This allows us to update the semantic probability vector $\bs_i $ recursively: 
% $
% \bs_i \leftarrow \frac{1}{Z} \bs_i \times \bs_i^t,
% $
In practice, this update is done in log space after a Laplace smoothing of $\bs_i^t$ for numerical stability. %The final class probability $\bs_i$ can be used for various tasks. For instance, one can extract its most likely class as the final semantic label $\pi_i = \arg\max_k \bs_{ik}$ and use it for navigation~\cite{asgharivaskasi2021active}, motion planning~\cite{9981300}, or exploration planning~\cite{ramakrishnan2022poni}.
Although other fusion methods exist (Sec.~\ref{sec:alternative fusion strategies}), RBU remains the most widely used thanks to its simplicity and probabilistic interpretation.

\begin{figure*}[htbp!]
    \centering
    \begin{subfigure}[t]{0.3\textwidth}
        \centering
        \includegraphics[width=.95\textwidth]{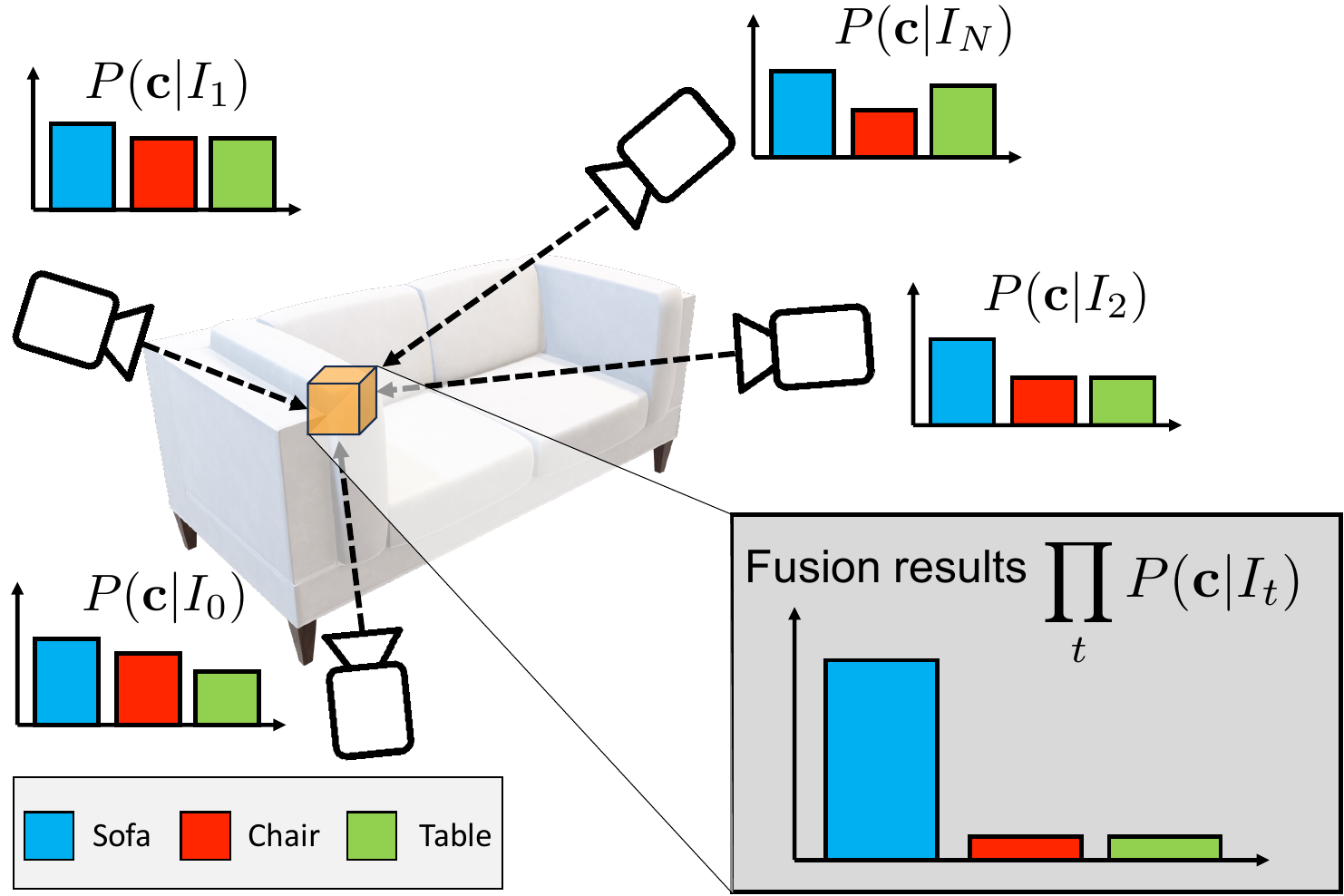}
        \caption{Aggregation overconfidence}
        \label{fig:bound_to_overconfidence}
    \end{subfigure}
    \begin{subfigure}[t]{0.3\textwidth}
        \centering
        \includegraphics[width=.95\textwidth]{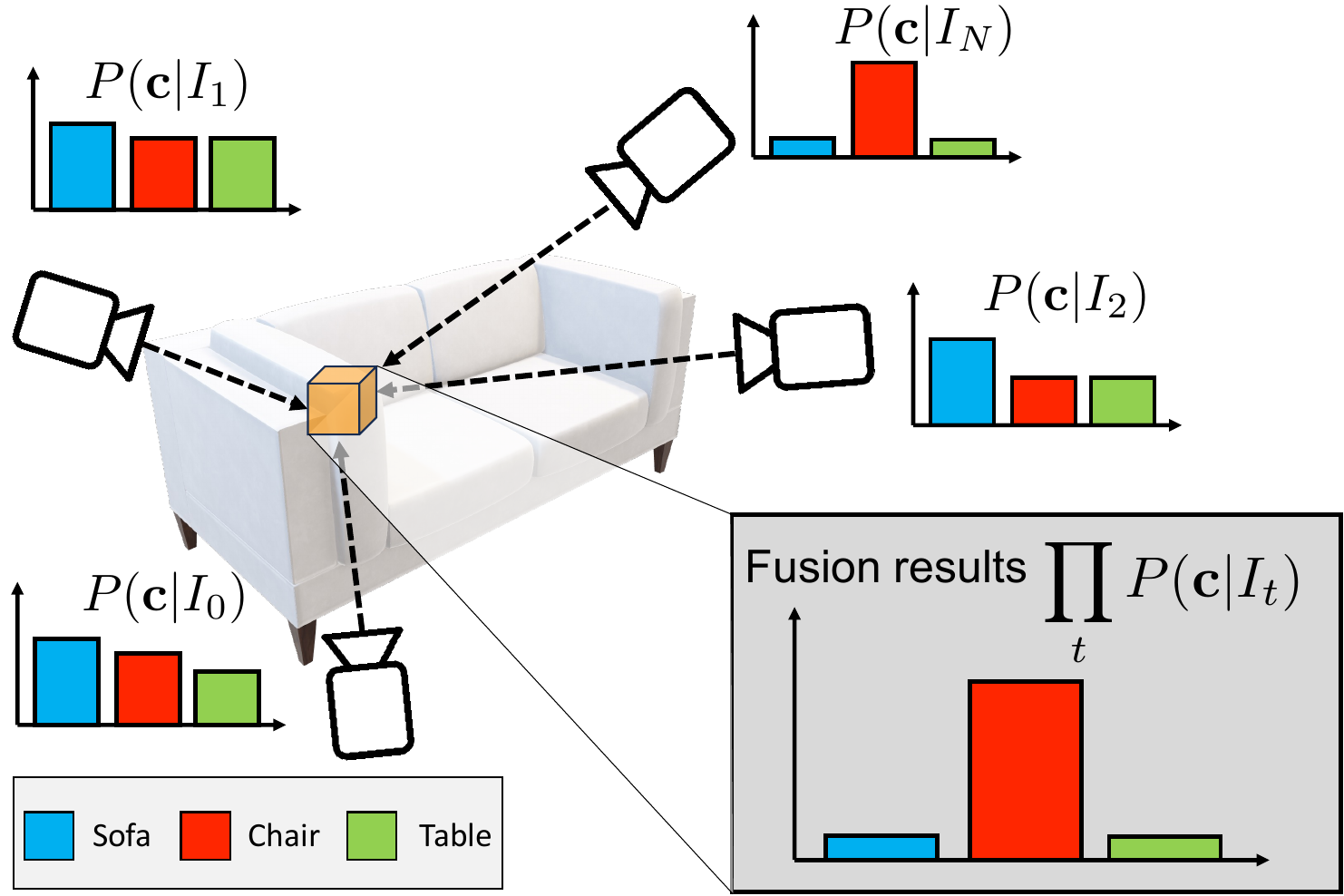}
        \caption{Sensitivity to outliers}
        \label{fig:outlier_sensitivity}
    \end{subfigure}
    \begin{subfigure}[t]{0.3\textwidth}
        \centering
        \includegraphics[width=.95\textwidth]{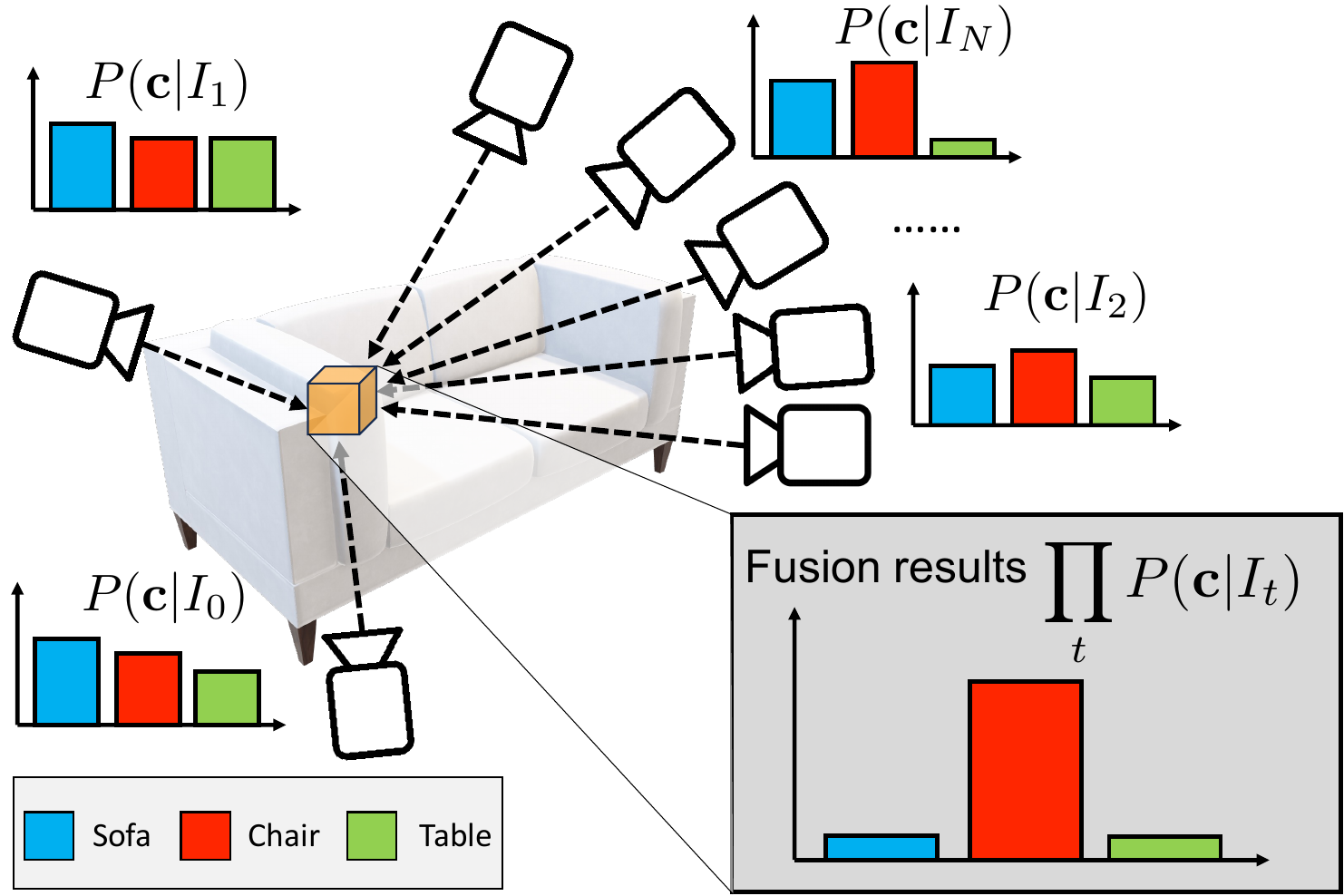}
        \caption{Viewpoint sampling biases}

        \label{fig:sampling_density_bias}

    \end{subfigure}
    \vspace{-5px}
    \caption{Illustrating potential causes of fusion overconfidence.  [Best viewed in color.]}
    \label{fig:Fusion Oveconfidence Origins}
    \vspace{-20px}
\end{figure*}

\subsection{Measuring Map Calibration}
\label{sec:Measuring map calibration}

We now discuss and present a few ways to measure 3D mapping miscalibration.
Let the ground truth label of a voxel $\bv_i$ be $l^*_i$, and its estimated posterior of belonging to class k, $\bs_{ik}$. Let $\pi_i$ denote the top-1 label of voxel $\bv_i$, $\pi_i = \arg\max_k(\bs_{ik})$, and  $h_i$ denote its predicted confidence, $h_i = \max_k(\bs_{ik})$.

The most widely used metric is Expected Calibration Error (ECE)~\cite{naeini2015obtaining,pmlr-v70-guo17a}. Partition the interval [0,1] in $O$ uniformly-spaced bins $B_b$ and match each sample to a bin by its $h(\bv_i)$ value. Define bin {\em confidence} as the mean value of the model's confidence for samples within the bin $\texttt{Conf}(B_b)=\frac{\sum_{h_i\in B_b}h_i}{|B_b|}$ and define the mean bin accuracy $\texttt{Acc}(B_b) = \frac{\sum_{h_i\in B_b}\mathds{1}[\pi_i = l^*_i]}{|B_b|}$.
The ECE is then the weighted difference between bin accuracy and confidence, $\texttt{ECE} = \sum_{b = 1}^O\ \frac{|B_b|}{N}\left| \Delta({B_b}) \right|$ where $\Delta(\cdot) = \texttt{Acc}(\cdot) - \texttt{Conf}(\cdot)$.

ECE is ill-suited for evaluating multi-class calibration since it is skewed by performance on common classes~\cite{Nixon2019MeasuringLearning,gupta2022toplabel}, and in 3D mapping, classes may have orders of magnitude more voxels than others. Many alternatives have been proposed to address these flaws \cite{Nixon2019MeasuringLearning,gupta2022toplabel,NEURIPS2019_8ca01ea9}.
 %It also only considers about the most likely prediction and it's measure is dependent on the size and number of the bins, being potentially manipulable, 
%Many alternatives have been proposed to address some of these issues, like Static Calibration Error \cite{Nixon2019MeasuringLearning}, which measures the calibration of all predictions and Adaptive Calibration Error, which controls bin size for ensuring near-equal bin membership. More relevant to our case, was 
Top-Label ECE (TL-ECE) \cite{gupta2022toplabel} performs binning conditioned on predicted class. It associates a sample with bin $B_{bk}$ if $h(\bv_{i}) \in B_b$ and $\pi_i = k$, and calculates a bin-weighted error:
% \begin{equation}
%     \Delta_{B_b} = \sum_{k = 1}^M\frac{|B_{bk}|}{\sum_{j=1}^{K} \left|B_{bj}\right|}\delta_{B_{bk}}
% \end{equation}
% with $\delta_{B_{lk}} = \left|Conf(B_{lk})-Acc(B_{lk})\right|$.
\begin{equation}
    \texttt{TL-ECE}
    = \sum_{b=1}^O \sum_{k = 1}^K \frac{|B_{bk}|}{N}|\Delta(B_{bk})|
\end{equation}

However, this metric is still heavily influenced by class frequency, since it weighs contributions to the metric by $\frac{|B_{bk}|}{N}|$.  We provide in this paper an alternative measure of calibration, which is agnostic to class frequency, mean Expected Calibration Error (mECE). Similarly to Class-Conditional ECE \cite{NEURIPS2019_8ca01ea9}, it is obtained by calculating a class-conditional ECE for every class, but unlike \citet{NEURIPS2019_8ca01ea9}, that bin predictions based on $h(\bv_i)$ and $\pi_i$, we bin predictions based on $h(\bv_i)$ and the {\em ground truth class}, $l*$, and take the average value across classes. This avoids the calibration failure case of a predictor obtaining perfect calibration by always predicting the most common class with the confidence given by its frequency. Thus, binning $B_{bk^*}$ is conditioned on $l^*=k$, and mECE is defined as:  
\begin{equation}
    \texttt{mECE} = \frac{1}{K}\sum_{k^* = 1}^K \sum_{b=1}^O\frac{|B_{bk^*}|}{\sum_{b=1}^O|B_{bk^*}|}|\Delta(B_{bk^*})|,
    \label{eq:mECE definition}
\end{equation}
retaining the notion of $\Delta(\cdot)$ from before.

We argue that this metric better captures the overall behavior of the 3D calibration model across all relevant classes, since it weights classes equally, i.e., %class-confidence bin contributions are only scaled by $\frac{|B_{lk}|}{\sum_{l=1}^L|B_{lk}|}$, i.e. 
miscalibration of a class' bins is scaled relative to their own class's frequency. The high-level difference between mECE and TL-ECE is analogous to the difference between mIoU and f-mIoU - i.e. mECE avoids drowning out minority classes.% when compared to TL-ECE. %Similarly, mECE can be taken over different sets of classes, such as "things" and "stuff", or over all classes, which we call mECE-T, mECE-S and mECE-A, respectively, which are also reported. - omitted here since we don't report these values in the end for brevity

\section{THE FUSION OVERCONFIDENCE PROBLEM}
\vspace{-5px}
\textbf{Accuracy} and \textbf{calibration} are the two principal desiderata for semantic 3D mapping. If we cannot guarantee 100\% accuracy, we at least expect our predictions to be well-calibrated. %That is, the confidence scores in our predictions should align closely with our expected performance for a given prediction \cite{NEURIPS2019_f8c0c968}.
For instance, if we predict that a voxel is a sofa with a 60\% probability, denoted as \( s_{ik} = 0.60 \), then this prediction should be correct about 60\% of the time. If a model consistently gives confidence scores that exceed its actual performance, it is deemed \textbf{overconfident}. Such overconfidence is often observed in 2D image-based perception tasks, and various uncertainty calibration strategies have been employed to address it \cite{NEURIPS2019_8ca01ea9,gupta2022toplabel,kuppers2022confidence}.

However, calibration in 3D semantic mapping remains unaddressed. As briefly noted by prior work~\cite{8490953}, The Recursive Bayesian Update ~\cite{7989538} in Eq.~\ref{eq:bayes_integration} tends to produce highly overconfident 3D semantic maps. We further illustrate this phenomenon in Fig.~\ref{fig:Fusion Overconfidence Problem}. %In it, we use fusion to integrate various semantic maps $\mathbf{S}^t$ %, ensuring that the uncertainty of each semantic map is precisely quantified using the calibration strategy described in  \todocite. 
Note that despite the fact that the 2D semantic cues exhibit only mild miscalibration, the resulting fused map using RBU displays severe overconfidence or worse disparity between accuracy and confidence.

Fusion overconfidence can have a few different causes, some of which are illustrated in \autoref{fig:Fusion Oveconfidence Origins}. 
Specifically, some of these reasons are:

\noindent
\textbf{Uncalibrated 2D Segmentation}: Many image-based semantic segmentation models are poorly calibrated \cite{kuppers2022confidence}. In particular, models trained with cross-entropy loss often demonstrate overconfident behavior \cite{pmlr-v70-guo17a}. Integrating overconfident estimates with RBU will result in an overconfident 3D prediction due to its sensitivity to outliers. % \autoref{fig:outlier_sensitivity}% \textcolor{red}{TODO: describe why.}

\noindent
\textbf{Incorrect Independence Assumption}: 
RBU assumes semantic likelihood observations are independent, overlooking dependencies across views and 2D model biases. Such biases exist in 2D deep models due to a variety of reasons, such as dataset collection biases \cite{kuppers2022confidence}, leading to the ``double-counting'' effect, as depicted in \autoref{fig:bound_to_overconfidence}. For instance, a calibrated model might consistently predict $\mathbf{s} = (0.49, 0.51)$ across 50 nearby viewpoints. RBU predicts class 2 with 88\% confidence, despite each observation being mostly noise. 

\noindent
\textbf{Sensitivity to Outliers}: As pointed by \citet{Morilla-Cabello2023RobustMapping}, a single overconfident input can drastically change the RBU posterior, as in \autoref{fig:outlier_sensitivity}. Laplace Smoothing of semantic likelihoods mitigates but does not fully solve this.

\noindent
\textbf{Viewpoint Coverage Dependence -} Most data collection in the real world cannot guarantee a well-covered, uniform sampling of viewpoints. This inevitably results in a lack of dense coverage, e.g., one might only see the sofa from a certain angle. If deep models are sensitive to viewpoints \cite{kuppers2022confidence}, as in \autoref{fig:sampling_density_bias}, this can lead to confident errors.

\noindent
\textbf{Non-Uniform Viewpoint Density -} Similarly, if the camera trajectory lingers longer in one region than another, as shown in \autoref{fig:sampling_density_bias}, positional biases in the model can steer it towards overconfidence, despite good pose diversity.

\noindent    
\textbf{Metric Mapping Errors -} Errors in geometric map reconstruction, whether from depth sensing errors or uncertainties in metric reconstruction, can result in spurious 2D-3D associations, leading to overconfident or inconsitent predictions.

Notably, 2D overconfidence is only a small contributor to fusion overconfidence. We calibrate two semantic segmentation models: Segformer ~\cite{https://doi.org/10.48550/arxiv.2105.15203} and ESANET~\cite{9561675} with Temperature Scaling on 21 scenes from ScanNet~\cite{Dai_2017_CVPR}. We then reconstruct another set of 100 scenes using RBU fusion for both uncalibrated and calibrated models. \autoref{tab: 2D calibration is insufficient} shows the mECE of the RGBD image pixels and the mECE of the voxels of the fused map. Traditional 2D calibration improves \textbf{pixel} mECE but \textbf{voxel} mECE is barely changed.  In contrast, we will introduce a 3D temperature scaling method for calibrating voxel mECE directly, leading to maps that are much better calibrated. Surprisingly, segmentation models that are calibrated to yield better calibrated maps can be quite uncalibrated at the pixel level!
\vspace{-3px}
%Other 2D post-hoc calibration methods exist \cite{NEURIPS2019_8ca01ea9,naeini2015obtaining,NEURIPS2019_f8c0c968,gupta2022toplabel}, but studying them and their 3D extension is left to future work.

\begin{table}[tbp]
    \centering
    \caption{{\bf Quantitative Analysis of Logit Scaling}: 2D calibration is insufficient to produce well-calibrated 3D semantic maps.}
    \label{tab: 2D calibration is insufficient}
    \resizebox{0.8\columnwidth}{!}{
    \begin{tabular}{llcc}
        \textbf{Model} & \textbf{Calibration} & \textbf{Pixel mECE} $\downarrow$ & \textbf{Voxel mECE}$\downarrow$ \\ \hline
        \multirow{3}{*}{\textbf{Segformer}}& None & 0.124 & 0.451 \\ 
        & 2D Temperature & \textbf{0.114} & 0.450 \\ 
        & 3D Temperature & 0.546 & \textbf{0.176}
 \\ \hline
        \multirow{3}{*}{\textbf{ESANet}}& None & 0.187 & 0.292
 \\ 
        & 2D Temperature & \textbf{0.159} & 0.294 \\ 
        & 3D Temperature & 0.656 & \textbf{0.177} \\ 
    \end{tabular}

    }
    \vspace{-20px}
\end{table}

\section{METHODS}
\vspace{-3px}
We identify and discuss three classes of approaches to address fusion overconfidence in the context of real-time semantic mapping: \textbf{alternative fusion strategies}, \textbf{down-weighting or filtering samples} and \textbf{calibrating the semantic segmentation model.}
% \begin{itemize}
    % \item 
    % \item .
    % \item 
% \end{itemize}
Finally, we also present a generalized method that simultaneously learns the fusion, weighting, and image segmentation calibration to optimize 3D calibration metrics without harming the accuracy. 
% and discuss different suitable metrics for 3D map calibration. % we already discussed in background

\vspace{-3px}
\subsection{Other Fusion Strategies}
\label{sec:alternative fusion strategies}
\vspace{-3px}
Alternative fusion strategies have been proposed to address some of the limitations of RBU. The \textbf{Histogram} fusion approach considers each image label as a ``semantic vote'' in favor of a class. The element probability is then given by:
\vspace{-10px}
\begin{equation}
    \bs_i \propto \frac{1}{T}\sum_{t=1}^T\mathds{1}[k = \arg \max s_i^t]
    \label{eq:histogram_accumulation_equation}
\end{equation}
(For brevity, treat the voxel as being visible at all $t=1,\ldots,T$.)
This strategy is widely used in ObjectNav \cite{chaplot2020object,ramakrishnan2022poni}, panoptic segmentation \cite{8741085} and outdoor reconstruction \cite{7138983}.

The \textbf{Na\"ive Averaging} idea treats the image segmentation likelihoods as fractional votes for each class~\cite{8490953,7138983,9811877}:
\begin{equation}
    \bs_i \propto \frac{1}{T}\sum_{t=1}^T \bs_i^t
    \label{eq:naive_averaging_equation}
    \vspace{-5px}
\end{equation}
Both histogram voting and Averaging voting integrate the semantic logits additively, which compromises their probabilistic interpretation. However, they are less sensitive to overconfident prediction outliers.
% \shenlong{need one sentence to justify why averaging is considered, same as histogram}
%In the context of naive averaging: $\eta_t^i$ is a per-voxel, per image weight attributed to each sample, either set to one or dependent on its distance to the camera and other heuristics\textcolor{red}{KH: what other heuristics have been used in prior approaches? Can we split this into a weighting section} \cite{8490953,7138983,9811877}. 

A final alternative we consider is \textbf{Geometric Mean} of likelihoods, as a logical extension of Na\"ive Averaging% \textcolor{red}{KH: citations?}:
\begin{equation}
    \bs_i \propto \left(\prod_{t=1}^T \bs_i^t\right)^\frac{1}{T},
    \label{eq:Geometric Mean}
    \vspace{-3px}
\end{equation}
which is also performed in log space after Laplace smoothing of $\bs_i^t$ for numeric stability. The geometric mean has the advantage of being less sensitive to outliers and is less prone to "double-counting" compared to RBU. 

Other alternatives exist, like max-fusion \cite{zhang20223d} and deep learned fusion~\cite{xiang2017rnn,9636599}, but are likely to be more susceptible to outliers \cite{zhang20233d} or not real-time capable\cite{xiang2017rnn,9636599}.

\subsection{Sample Weighting}

Since semantic segmentation models and depth sensors can exhibit distance and pose related biases \cite{kuppers2022confidence}, one might not wish to give all observations equal weights in fusion \cite{curless1996volumetric}. As such, each of the integration strategies above can be modified to use weights to capture some degree of confidence in each pixel's contribution to the 3D element {\em in dimensions orthogonal to the image segmentation confidence}. For example, \citet{curless1996volumetric} propose to weigh samples based on the estimated normal and distance to the camera, \citet{kuppers2022confidence} proposes image-position-based calibration of uncertainties, and \citet{9981300} propose a quadratic sample weight based on voxel distance to camera to reflect a "segmentation distance sweet-spot". 
Recent work proposes to weigh samples based on estimated epistemic uncertainty using Monte-Carlo dropout on segmentation networks \cite{Morilla-Cabello2023RobustMapping}, which is ill-suited for real-time applications.

\subsection{Calibrating Image Segmentation via Logit Scaling}

The final category of approaches we consider is to calibrate the confidence in image segmentation likelihoods $\bs_i^t$. In this paper, we consider two baseline calibration methods, Temperature and Vector Scaling, known to work well in 2D semantic tasks \cite{pmlr-v70-guo17a}. Let the image semantic likelihoods be $\mathbf{S}^t = \sigma_{SM}(\mathbb{\lambda}^t)$, where $\mathbb{\lambda}^t$ are the segmentation model logits and $\sigma_{SM}$ is the softmax function. Define a temperature  parameter $\tau$ and a calibrated likelihood as $\mathbf{S}^t(\tau) = \sigma_{SM} \left(\frac{\mathbb{\lambda}^t}{\tau}\right)$. Given a calibration metric $\Omega$ such as mECE, 2D temperature scaling minimizes calibration error of all images over $\tau$:
\begin{equation}
    \tau = \arg\min_{\tau}\Omega\left( \{ \mathbf{S}_{u,v}^t(\tau) \quad\forall u,v,t\} \right).
    \vspace{-5px}
\end{equation}
Vector scaling uses a per-class scaling vector $\tau\in\mathbb{R}^K$ and scales the logits classwise as $\mathbf{S}_{uv}^t(\mathbf{\tau}) = \sigma_{SM} \left(\left\{ \frac{ {\mathbb{\lambda}_{u,v,k}^t}}{\tau_k} \quad\forall k=1,\ldots,K \right\} \right)$. %Zero-th order optimization methods, e.g., Bayesian Optimization \cite{NogueiraBayesianPython}, are typically used. 

This paper introduces a new concept of {\em 3D Temperature / Vector scaling}. The calibration metric is no longer assessed over images but rather over the fused voxels of the map. For each voxel $\bv_i$, its estimated posterior is a function of $\tau$, $\bs_i(\tau) = \texttt{Fusion}(\bs_i^1(\tau),\ldots,\bs_i^T(\tau))$, where $\texttt{Fusion}(\ldots)$ is any of the described methods in Section~\ref{sec:alternative fusion strategies}. This can then similarly be optimized in vector and temperature variants:
\vspace{-5px}
\begin{equation}
    \tau = \arg\min_{\tau}\Omega(\{ \bs_i(\tau) \quad \forall \bv_i\in \bV \})
    \label{eq:vector_3d_calibration}
\end{equation}
Logit scaling is usually performed with multiple image sequences, seeking to minimize the mean calibration error.
\vspace{-5px}

% \textcolor{red}{TODO: describe 3D temperature scaling and vector scaling}

% Performing calibration on semantic fusion pipelines for 3D reconstruction is not trivial. Since the output of the calibration is only obtained after the fusion equations are applied, i.e., Eqs. 4 - 8 - and common calibration methods, like temperature and vector scaling \cite{pmlr-v70-guo17a} have non-linear effects on the fused probits, there is no immediate way of bypassing the need for performing fusion at every evaluation. Let $\sigma_{sm}$ denote the Softmax function and $l_i$ denote some model's logits, note how $\Pi_{i=0}^N\sigma(l_i/T)\neq \Pi_{i=0}^N\sigma(l_i)/T$ or $\Sigma_{i=0}^N\sigma(l_i/T)\neq \Sigma_{i=0}^N\sigma(l_i)/T$. Thus, to enable expedient optimization, we should cache the model logits for every input in the reconstruction system, which requires significant storage (or compute them on the fly using smart buffering, at the cost of GPU cycles). Thus, even for a small set of calibration validation scenes and for a real-time capable system, reconstruction can take several minutes per scene, depending on the number of frames of the recording, resolution and storage performance to load the cached logits. 

\subsection{Generalized Learned Fusion Strategy (GLFS)}

Finding the optimal mix of fusion, sampling, and logit scaling strategies is cumbersome. We address this challenge by introducing a generalized learned fusion strategy, GLFS. 
The key idea is to use differentiable gating variables to switch between various integration strategies and to make both the temperature and sample weighting learnable as in Eq. \ref{eq:generalized fusion definition}%
\vspace{-10px}
% Let $\mathbf{l}_i^t \in \mathbb{R}^M$ denote a logit observation of associated with voxel $v_i$ at time $t$. Let $\mathbf{\tau} \in \mathbb{R}^{+M}$ be a learned temperature vector and $\eta = | \bm{\tau} |^2$ be its L2-norm. Let $w_i^t$ be a learned sample weight associated with each observation of voxel $\bv_i$, $t=1,...,T$ and let $G,\epsilon \in \mathbf{R}^{[0,1]}$ be two extra learned parameters. The generalized fused unnormalized probability, $\tilde{\bs_i}$ for this voxel at time T is then given  \autoref{eq:generalized fusion definition}, where $\alpha_T$  where $ \alpha_T = \left(\frac{1-\epsilon}{\Sigma_{t=1}^Tw_i^t} + \epsilon\right)$
%    \label{eq:alpha_t_def}$ and $\mathbf{\tilde{p}}_i^t = P(l_i,\mathbb{\tau}|\bI_t)$.

\begin{equation}
   \bs_i \propto G~e^{\left({\Sigma_{t}\alpha\left(w_i^tln(\bs_i^t(\tau))\right)}\right)} + (1-G)\Sigma_{t}\alpha\left(w_i^t\bs_i^t(\tau)\right)
   \label{eq:generalized fusion definition}
   \vspace{-5px}
\end{equation}
where $\alpha = \left( \frac{(1-\epsilon)}{\Sigma_{t}w_i^t} + \epsilon\right)$ is a gating variable that interpolates between RBU and Geometric Mean, while $G$ is a gating variable that balances between the arithmetic mean and the geometric mean for integration. $w_i^t$ weighs the importance of each pixel contributing to the final voxel semantics and $\tau$ represents the logit scaling temperature.

% Normalizing $\tilde{\bs}_i$ to sum to 1 provides an estimate of the fused posterior, $\hat{\bs}_i$. \shenlong{why using tilde and hat? wasn't mentioned before}
This unified procedure generalizes to all the listed fusion strategies and incorporates logit adjustments via vector scaling and pose-dependence adjustments through learned sample weighting. When $\tau = 1$ and $w_i^t = 1$ $\forall i, t$, it reduces to RBU, Geometric Mean, and Na\"ive Averaging when $(G,\epsilon) = (1,1)$, $(1,0)$, $(0,0)$, respectively. Similarly, it reduces to histogram fusion when $|\tau| \rightarrow 0$ and $(G,\epsilon) = (0,0)$. Thus, this model is capable of capturing the behavior of all fusion strategies we have detailed. 

To account for pose biases in segmentation models, $w_i^t$ comes from a learned look-up-table, $\bm{M}$, which takes as entries $\pi_i^t$ (i.e, the post-scaling predicted class of the pixel associated with this voxel at this time), $d_i^t$, the distance from this voxel to the camera, and $\alpha_t^i$, the incidence angle between the camera ray and the current estimate of the surface normal at that voxel, hopefully capturing the effects of the weighing heuristics through learning.  We define $\theta=(\tau,G,\epsilon,\mathbf{M})$, and the
calibration error is defined as  $\texttt{mECE}(\{ \mathbf{s}_i(\theta) \,\forall i\})$.

However, mECE is not directly end-to-end trainable because it is not differentiable, so we use a differentiable analogue, DECE \cite{Bohdal2021Meta-Calibration:Error}, for which we can also define its mECE equivalent, which we call mDECE. 
Our overall loss function balances calibration with accuracy:
\vspace{-5px}
\begin{equation}
    \label{eq:Loss function}
    \mathcal{L} = \eta~\texttt{mDECE}(\bs_i(\theta),l_i^*) + \texttt{NLL}(\bs_i(\theta),li^*)
    \vspace{-5px}
\end{equation}
where $\eta$ scales the calibration term relative to the accuracy term (Negative Log Likelihood). 
This loss simultaneously promotes uncertainty calibration while minimizing the compromise in prediction accuracy. 
The calibration model parameters are then learned through backpropagation with PyTorch \cite{NEURIPS2019_9015}, leveraging voxel caching for efficiency.

\section{EXPERIMENTS}

\subsection{Segmentation Models and Datasets}

To evaluate the effect of different % fusion strategies and calibration methods 
strategies on 3D fusion calibration, we perform reconstruction experiments on the ScanNet Dataset \cite{Dai_2017_CVPR} using two different semantic segmentation models: Segformer~\cite{https://doi.org/10.48550/arxiv.2105.15203} and ESANet~\cite{9561675}. 
These are representative segmentation networks for RGB and RGB-D, respectively. Moreover, they utilize two differing backbone architectures: Vision Transformers \cite{dosovitskiy2020vit}
and Convolutional Neural Networks, \cite{He_2016_CVPR}.
to help us evaluate the generality of the various calibration strategies.
% These models were chosen because they represent both two different segmentation modalities, RGB and RGB-D, as well as two common architectures for segmentation, Vision Transformers and Convolutional Neural Networks. 
Segformer was pretrained on the ADE-20K dataset using the b4 backbone \cite{Https://huggingface.co/nvidia/segformer-b4-finetuned-ade-512-512} and ESANet was pretrained on the NYU\_v2 dataset \cite{Silberman:ECCV12}. Both models were then finetuned on ScanNet's 20 classes on a subset of every 30-th frame of the train split recordings of the ScanNet dataset, with their classification head trained from scratch.  We implement mapping strategies using Open3D's real-time GPU-based dense SLAM implementation \cite{9918017}.

We divided the scans from the ScanNet V2 validation set into calibration training (79 scans), calibration validation (21 scans), and calibration testing (100 scans). Since there are multiple scans for the same scene, we make sure there is no scene overlap between testing and validation/training sets.

% For evaluating base model calibration and calibration procedure efficacy, we divide Scannet V2's validation dataset into two groups of 100 disjoint scans, such that neither group contains a scene from the other group. One set is the \textbf{calibration test}, while the other one is the \textbf{calibration tuning} set. The calibration tuning set is similarly subdivided in one set of 21 and 79 scans, such that the set of 21 scans contains at least 2 instances of each of the named classes and shares no scenes with the 79 scans one. The smaller set is used as \textbf{calibration validation} set, while the larger one is used for \textbf{calibration learning} of the end-to-end model.

%We first sample 100 scans from the validation set to be our \textbf{test set}. Then, we remove the scenes to which these scans belong from the pool and, sample 100 scans to be our \textbf{"calibration" set}. The calibration set is then further subdivided into a \textbf{"calibration training" set} and a \textbf{"calibration validation"} set by randomly sampling sets of 20 scenes until we have at least 3 instances of each of the 20 classes in our sample. We then take all of the scans in this set which contain that scene - and reserve the resulting 21 scans as validation scans. The remaining 79 scans are then used to train our calibration model.

\begin{figure*}[htbp!]
    \centering
    \includegraphics[width=0.7\linewidth]{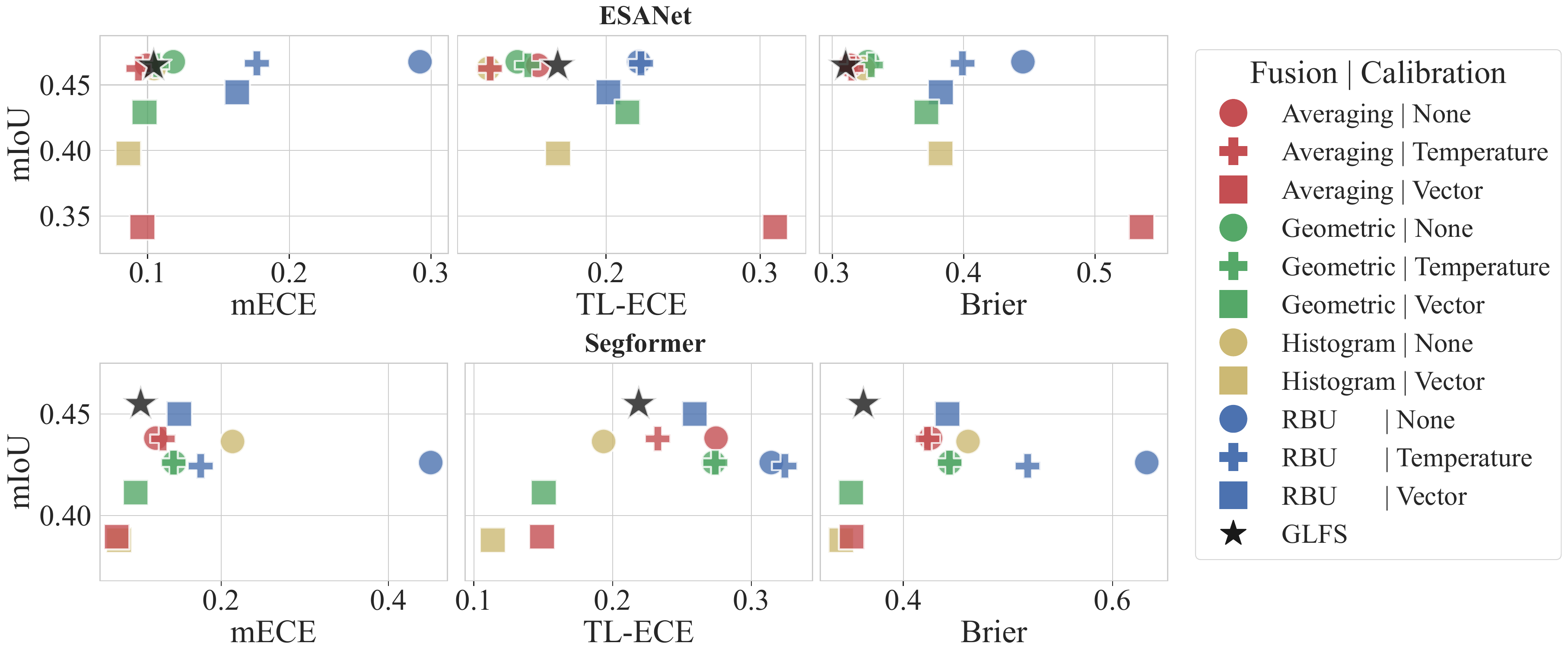}
    \vspace{-5px}
    \caption{Map calibration vs. accuracy for combinations of fusion and calibration strategies on 100 ScanNet scenes. Higher is better for mIOU; lower is better for mECE, TL-ECE, and Brier - i.e. better models are always closer to the upper left corner. [Best viewed in color.]}
    \label{fig:Calibration_differences_between_fusion_strategies}
    \vspace{-10px}
\end{figure*}

\subsection{Mapping Calibration Experiments}
\label{sec:mapping_calibration_experiments_section}

We compare the calibration and accuracy of the fusion strategies and 3D calibration methods described in Sec. III. %As baseline calibration strategies we consider both Temperature and Vector scaling \cite{pmlr-v70-guo17a}. These calibration parameters are optimized in 3D as defined by \autoref{eq:vector_3d_calibration} on the calibration validation scans, using voxel mECE as the optimization target. 

Bayesian optimization \cite{NogueiraBayesianPython} is used for temperature and vector scaling, using the mECE metric and Upper Confidence Bound (UCB) acquisition function with $\beta = 1$, the Mattern 2.5 kernel and 1000 maximum samples (or 48 hours, whichever came first). The models were initialized with a sweep of 50 temperatures in log space between the maximum and minimum temperatures for temperature calibration (0.01, 200). Then, we narrowed the search for the vector scaling parameters to be closer to the optimal temperature scaling parameter, allowing temperatures to be within a 50\% difference of each optimized value. A similar log-scale diagonal sweep is performed as initialization for the vector parameters - as well as 30 random samples.

\begin{figure*}[tbp!]
  \begin{minipage}[b]{0.48\textwidth}
    \includegraphics[width=0.95\linewidth]{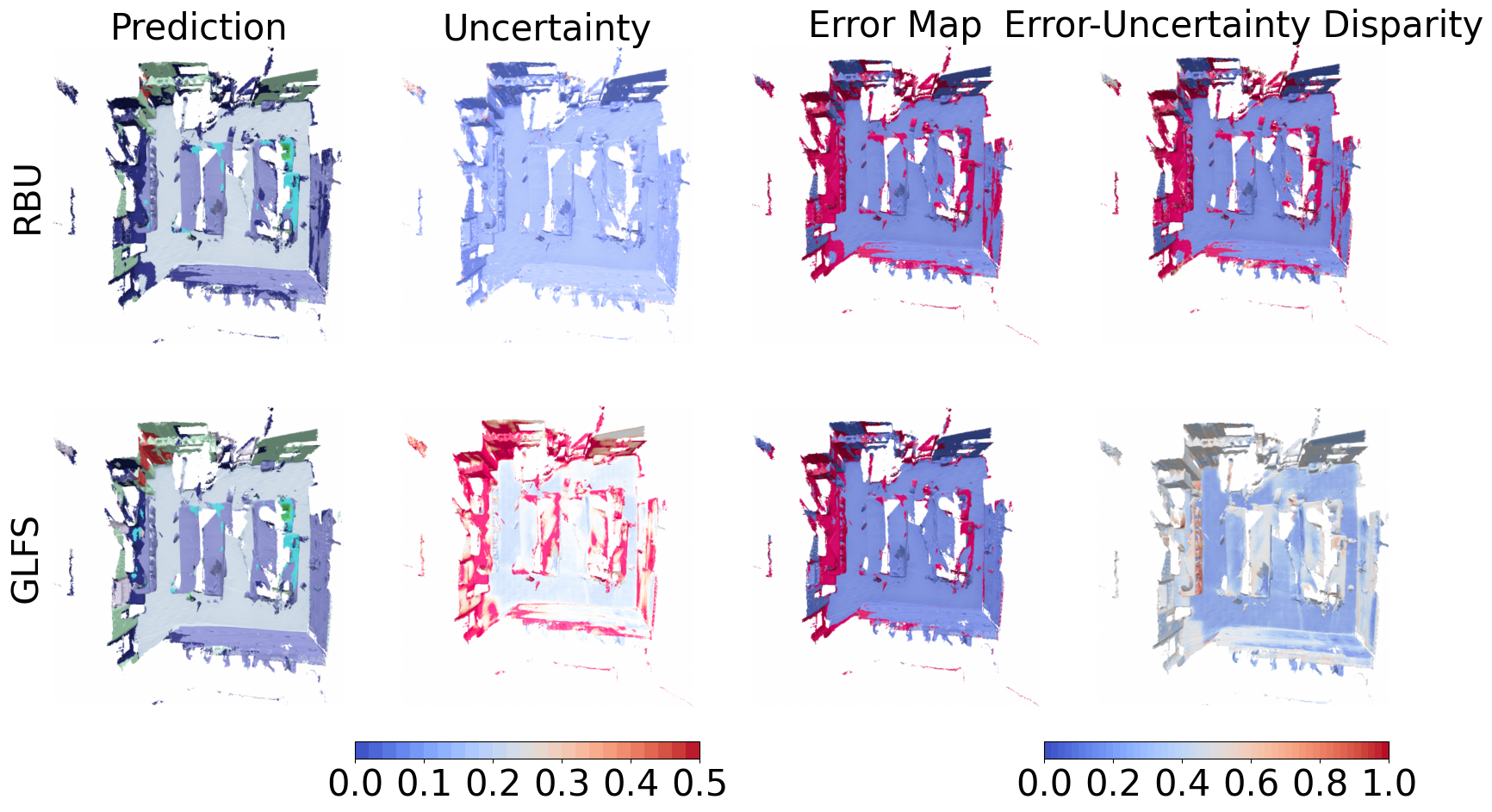}
    \label{fig:first_image}
  \end{minipage}
  \hfill
  \begin{minipage}[b]{0.48\textwidth}
    \includegraphics[width=0.95\linewidth]{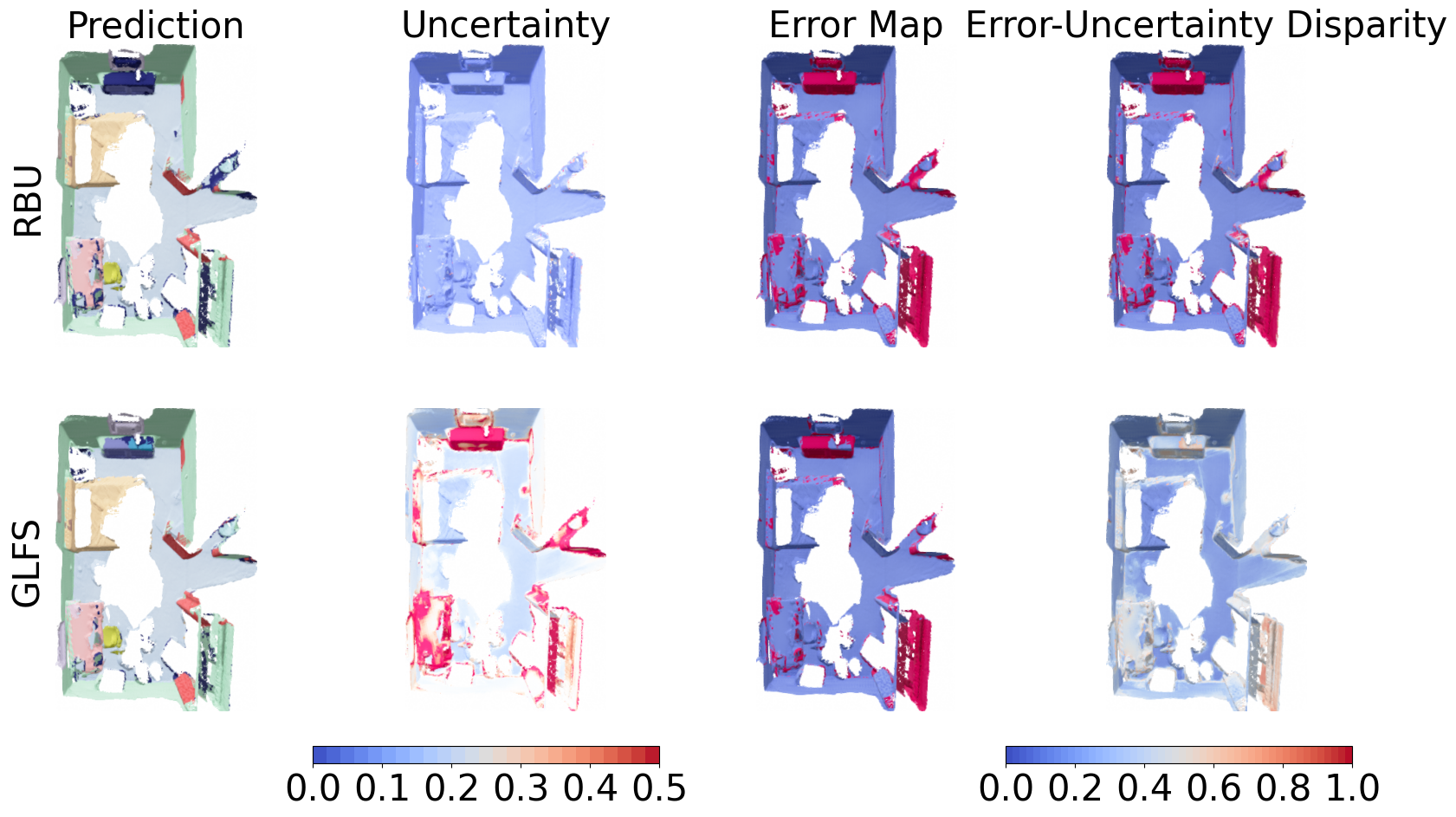}
    \label{fig:second_image}
  \end{minipage}
  \vspace{-2mm}
    \caption{Qualitative comparison of RBU (top) and our proposed GLFS method (bottom) on two ScanNet scenes reconstructed using the Segformer model. Although similarly accurate, GLFS provides significantly improved calibration. [Best viewed in color.]}
 \label{fig:qualitative example reconstruction improvement}
  \vspace{-20px}
 \end{figure*}

Results are reported on the calibration test split using ground-truth poses.  \autoref{fig:Calibration_differences_between_fusion_strategies} plots calibration (mECE, TL-ECE, and Brier Scores) vs accuracy (mIoU).  We observe a clear tradeoff between accuracy and calibration. Temperature scaling tends to somewhat improve calibration with minimal degradation in accuracy, but surprisingly, vector scaling is often highly detrimental to accuracy for sometimes modest calibration improvements. Averaging-based fusion is generally a strong performer. Our learned model typically exceeds other methods in either accuracy or calibration, improving calibration without degrading accuracy, as intended. We also found that the ESANet model proved harder to calibrate.

Qualitative results comparing RBU to our learned method are shown in Fig.~\ref{fig:qualitative example reconstruction improvement}. The error-uncertainty disparity for RBU is high, demonstrating severe overconfidence. In contrast, GLFS is less confident in its mistakes, improving mECE. 

    % Compute calibration of finetuned models using every fusion strategy, as well as calibration baselines (temperature and vector scaline), and finally the full learned pipeline, comparing them in terms of mIoU, accuracy, Brier Scores, mECE, ECE and TL-CE
    % This is done for both ESANet and Segformer to account for the two common semantic segmentation model types currently.

\vspace{-3px}
\subsection{Object Goal Navigation Experiments}
\vspace{-3px}
We integrate our semantic fusion pipelines with a recent modular Object Goal Navigation agent, PEANUT \cite{zhai2023peanut}. We use our live metric-semantic reconstruction pipeline to generate a 3D semantic point cloud with different fusion strategies, which is then projected into the 2D plane and overlaid over PEANUT's obstacle and exploration map. During the projection, points get binned based on their ground coordinates and on $\pi_i$ for each point. Then, the average of the point confidence predictions within that $(x,y,l)$ bin represents the confidence of the class $l$ being in $(x,y)$ on the 2D map. 

This semantically fused map is fed to PEANUT's goal prediction network \textbf{without finetuning it} to the new distribution of maps and semantics due to time constraints. Besides the semantic fusion, the only major difference between our version and PEANUT is that PEANUT's detection of an object is performed by thresholding an image-level confidence through Mask-RCNN % model \cite{He_2017_ICCV},
whereas we perform detection by thresholding the fused 2D semantic map level,i.e., on the projected $\bs_i$.
Finally, we apply connected component analysis to the goal map, retaining only the largest goal connected component to account for voxel misprojections.

We finetune a Segformer \cite{https://doi.org/10.48550/arxiv.2105.15203} model on a set of training images from random exploration on HM3D v0.1 \cite{yadav2022habitatmatterport} as our agent's semantic segmentation. 
We evaluate the performance of the agent under 3 different conditions: On vanilla PEANUT, by thresholding Segformer's semantic mask in image space, and on our semantically fused implementation with both RBU(Fusion-RBU) and Na\"ive Averaging (Fusion-NA) fusion. 
For each mixed model, we tune PEANUT's hyperparameters, like collision radius and semantic detection threshold, on the first 500 validation episodes % of HM3d 0.1 \cite{yadav2022habitatmatterport} 
of the 2022 Habitat ObjectNav challenge, and evaluate it on episodes 500-999 of the same challenge.
ObjectNav metrics \cite{batra2020objectnav} of these agents and a ground truth segmentation ablation, presented in \autoref{tab:ObjectNavResultsSummary}, show that the empirically better calibrated 3D Na\"ive Averaging fusion can result in an increase in better agent success rates when compared to \citet{chaplot2020object}'s Neural deterministic fusion (NDF) or the overconfident RBU, at the expense of requiring more evidence to acquire goal targets, resulting in slightly worse SPL.
\vspace{-5px}

\begin{table}[tbp]
    \caption{Object goal navigation performance on HM3D-val for various variations of PEANUT and (standard error of the mean). The calibration strategies studied in this paper generally improve performance.}
    \label{tab:ObjectNavResultsSummary}
    \centering
    \resizebox{0.95\columnwidth}{!}{
    \begin{tabular}{llllll}
        \textbf{Map Type} & \textbf{Segmentation} & \textbf{SPL} $\uparrow$& \textbf{Success} $\uparrow$& \textbf{SSPL} $\uparrow$ & \textbf{DTG} $\downarrow$ \\
        \hline
        \textbf{NDF \cite{chaplot2020learning}}& \textbf{GT} & 0.384 (0.0132) & 0.724 (0.0200)  & 0.401 (0.0123)& 2.876 (0.2534)\\ 
        \hline 
        \textbf{NDF \cite{chaplot2020learning}} & 
        Segformer & 0.295 (0.0130) & 0.596 (0.0220) & 0.328 (0.0121) & 3.764 (0.2639)\\ 
        \textbf{Fusion-NA} & Segformer &0.312 (0.0132)& \textbf{0.604} (0.0219)& 0.347 (0.0120)& 3.576 (0.2589) \\ 
        \textbf{Fusion-RBU} & Segformer & \textbf{0.314} (0.0134) & 0.596 (0.0220) & \textbf{0.351} (0.0123) & \textbf{3.549} (0.2595)\\ 
    \end{tabular}
    }
    \vspace{-23px}
\end{table}

% \begin{table}[!ht]
%     \centering
%     \resizebox{0.95\columnwidth}{!}{
%     \begin{tabular}{cccccc}
%         \textbf{Strategy} & \textbf{Segmentation} & \textbf{SPL} & \textbf{Success} & \textbf{SSPL} & \textbf{DTG} \\ \hline
%         \textbf{NDF} & GT & 0.527 (0.0115) & 1.0 (0) & 0.523 (0.0115) & 0.041 (0.0011) \\ 
%         \hline
%         \textbf{NDF} & Segformer & 0.388 (0.0145) & 0.776 (0.0219) & 0.407 (0.0134) & 1.479 (0.1856) \\ 
%         \textbf{Fusion-NA} & Segformer & 0.414 (0.0142) & 0.796 (0.0212) & 0.432 (0.0128) & 1.353 (0.1787) \\ 
%         \textbf{Fusion-NB} & Segformer & 0.414 (0.0148) & 0.779 (0.0218) & 0.437 (0.0132) & 1.334 (0.1680) \\ 
%     \end{tabular}
% }
% \end{table}

\section{CONCLUSIONS AND FUTURE WORK}
\vspace{-3px}
We introduce the problem of semantic calibration in real-time capable metric-semantic mapping pipelines and show that better calibration can be achieved through an end-to-end unified approach to semantic fusion, GLFS, without sacrificing segmentation performance. We also show that better semantic fusion improves the performance of modular ObjectNav agents when using the same semantic segmentation model by providing robustness to outlying predictions.

Three avenues are considered for future work: First, we want to quantify the effects of other calibration strategies like position-dependent calibration \cite{kuppers2022confidence} and Dirichlet calibration \cite{NEURIPS2019_8ca01ea9} on 3D semantic map calibration, and novel 3D specific calibration procedures. Second, we wish to investigate how ObjectNav agents can better leverage 3D semantic uncertainty to guide their exploration \cite{asgharivaskasi2021active} and how that would affect their performance. Finally, we would like to study the quantification and improvement of uncertainty in open-set metric-semantic maps \cite{engelmannopen,jatavallabhula2023conceptfusion}.
% Other key takeaways: Naive Bayesian fusion has worse performance of all, despite being most commonly used, naive averaging or geometric averaging are recommended for practitioners
% Sample weighing based on distance incidence angle seems to have little impact on final performance (suggesting commonly used heuristics have small impact).

% \addtolength{\textheight}{0cm}   % This command serves to balance the column lengths
                
\newpage

\clearpage
%\bibliographystyle{IEEEtranSN}
%\bibliography{joao_references, other_references}

\printbibliography

\end{document}